\pdfoutput=1
\documentclass[11pt]{article}

\usepackage[]{acl}
\usepackage{times}
\usepackage{latexsym}
\usepackage[T1]{fontenc}
\usepackage[utf8]{inputenc}
\usepackage{microtype}

\usepackage{graphicx}
\usepackage{amsmath}
\usepackage{amssymb}
\usepackage{multirow}
\usepackage{array}
\usepackage{booktabs}
\usepackage{subcaption}
\newcommand\blfootnote[1]{
  \begingroup
  \renewcommand\thefootnote{}\footnote{#1}
  \addtocounter{footnote}{-1}
  \endgroup
}

\title{Shimo Lab at ``Discharge Me!'': Discharge Summarization by Prompt-Driven Concatenation of Electronic Health Record Sections
}

\author{
   Yunzhen He$^{\ast\,1}$ \qquad Hiroaki Yamagiwa$^{\ast\,1}$ \qquad Hidetoshi Shimodaira$^{1,2}$ \\
  $^1\,$Kyoto University \qquad $^2\,$RIKEN AIP\\
  \texttt{he.yunzhen.25d@st.kyoto-u.ac.jp,}\\
  \texttt{hiroaki.yamagiwa@sys.i.kyoto-u.ac.jp,shimo@i.kyoto-u.ac.jp}\\
}

\newcommand{\pink}[1]{\textcolor[rgb]{1.00,0.00,1.00}{#1}}

\begin{document}
\maketitle
\begin{abstract}
In this paper, we present our approach to the shared task ``Discharge Me!'' at the BioNLP Workshop 2024. The primary goal of this task is to reduce the time and effort clinicians spend on writing detailed notes in the electronic health record (EHR). Participants develop a pipeline to generate the ``Brief Hospital Course'' and ``Discharge Instructions'' sections from the EHR. Our approach involves a first step of extracting the relevant sections from the EHR. We then add explanatory prompts to these sections and concatenate them with separate tokens to create the input text. To train a text generation model, we perform LoRA fine-tuning on the ClinicalT5-large model. On the final test data, our approach achieved a ROUGE-1 score of $0.394$, which is comparable to the top solutions.
\blfootnote{$^\ast$ The first two authors contributed equally to this work.}
\blfootnote{Our code is available at \url{https://github.com/githubhyz/DischargeMe_BioNLP2024}.}
\end{abstract}

\section{Introduction}
Electronic health records (EHR) eliminate the need for end-users to write medical records by hand and provide easy access to digital records~\cite{Menachemi2011Benefits}.
However, the use of EHR sometimes increases the burden on end-users~\cite{Shanafelt2016Relationship,liu2022note,DBLP:conf/bionlp/GaoD0A23}. 
With this in mind, there has been active research in recent years into applying natural language processing (NLP) to EHR to reduce the burden on end-users~\cite{DBLP:journals/npjdm/0002FWAMJCW22,Houssein2023,VanVeen2023}.

\begin{figure}[t!]
    \centering
    \includegraphics[width=\columnwidth]{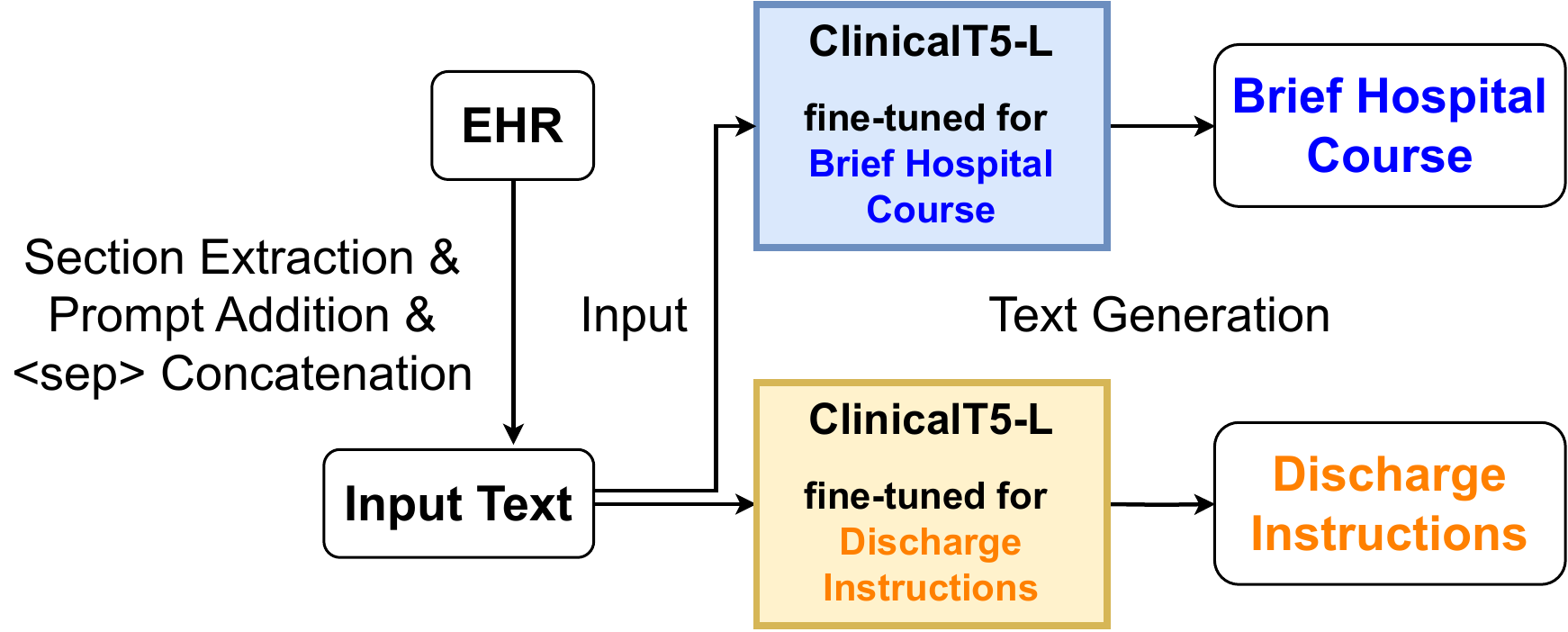}
    \caption{
Overview of our pipeline. To create input text, we extract sections from the EHR, add explanatory prompts, and then concatenate them with \texttt{<sep>} tokens. We then generate discharge summaries using ClnicalT5-large, which has been fine-tuned for each target.
}
    \label{fig:pipeline}
\end{figure}

To explore the potential of NLP in EHR, the shared task ``Discharge Me!''~\cite{xu-etal-2024-overview} at the BioNLP Workshop 2024 evaluates the ability to generate discharge summaries. 
The goal of this task is to reduce the time and effort clinicians spend on writing detailed notes in the EHR. 
Participants develop a pipeline that leverages the EHR data to generate discharge summaries.

\begin{figure}[t!]
    \centering
    \includegraphics[height=0.85\textheight]{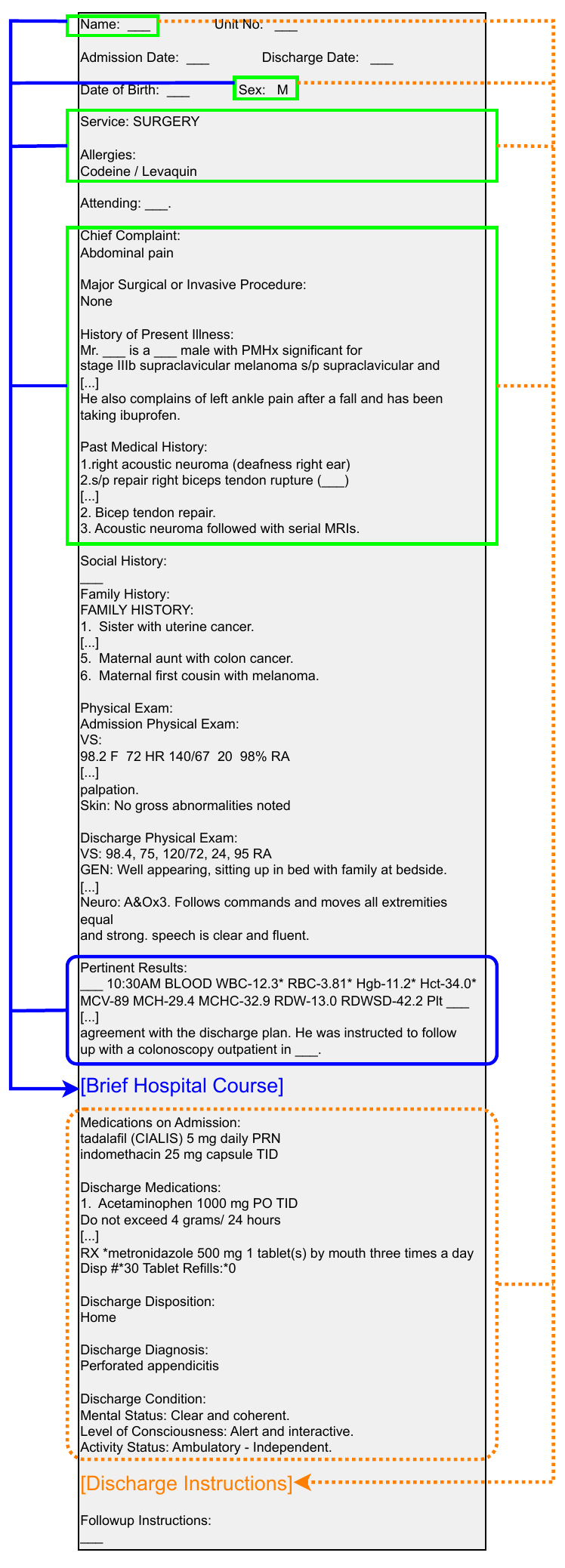}
    \caption{
An example of the EHR with the location of the target discharge summaries. To show the sections used for the input text, the {\color{blue}rounded rectangle} is for the ``{\color{blue}Brief Hospital Course}'', the {\color{orange}dashed rounded rectangle} is for the ``{\color{orange}Discharge Instructions}'',
and the {\color{green}rectangles} are for both targets.
The symbol ``[...]'' indicates omissions. 
}
\label{fig:ehr}
\end{figure}

In this paper, we present our approach to the shared task. 
Fig.~\ref{fig:pipeline} provides an overview of our pipeline. 
We preprocess the EHR, as illustrated in Fig.~\ref{fig:ehr}, by removing noise and extracting sections that are essential for the target summary. The sections are selected based on a predetermined priority. For extracted sections, we prepend the prompt from Table~\ref{tab:section-prompt} to the beginning of the text, concatenate these sections using \texttt{<sep>} tokens, and thus prepare the input text. 
We also removed noise from the target text. 
We then fine-tuned ClinicalT5~\cite{DBLP:conf/emnlp/LuDN22}, which is pre-trained on clinical texts.
On the final test data, our approach achieved a ROUGE-1 score of $0.394$, which is comparable to the top solutions.

\section{Related work}
\subsection{Text generation models in clinical domain}

\paragraph{Decoder.} 
 ClinicalGPT~\cite{DBLP:journals/corr/abs-2306-09968}, whose base model is BLOOM-7B~\cite{le2022bloom}, uses LoRA~\cite{DBLP:conf/iclr/HuSWALWWC22} for fine-tuning and applies the reinforcement learning process used in InstructGPT~\cite{DBLP:conf/nips/Ouyang0JAWMZASR22}. 
 BioMistral-7B~\cite{DBLP:journals/corr/abs-2402-10373} underwent additional pre-training of the Mistral-7B~\cite{DBLP:journals/corr/abs-2310-06825} model on PubMed Central~\cite{roberts2001pubmed} and showed good performance on the clinical knowledge QA task.

\paragraph{Encoder-decoder.} 
ClinicalT5~\cite{DBLP:conf/emnlp/LuDN22,DBLP:conf/chil/LehmanHMWSZNSJA23}, whose base model is T5~\cite{DBLP:journals/jmlr/RaffelSRLNMZLL20}, is the model pre-trained on clinical texts\footnote{Both of \citet{DBLP:conf/emnlp/LuDN22} and \citet{DBLP:conf/chil/LehmanHMWSZNSJA23} refer to their models as ClinicalT5.}. 
\citet{DBLP:conf/emnlp/LuDN22} performed additional pre-training of the SciFive-PubMed-PMC~\cite{DBLP:journals/corr/abs-2106-03598} model on MIMIC-III~\cite{johnson2016mimiciii}. Meanwhile, \citet{DBLP:conf/chil/LehmanHMWSZNSJA23} pre-trained T5 from scratch using MIMIC-III and MIMIC-IV~\cite{johnson2023mimiciv}.

\subsection{Clinical text summarization}
\paragraph{Discharge Summarization.}
\citet{williams2024evaluating} showed that although $33\%$ of the discharge summaries generated by GPT-4~\cite{achiam2023gpt} from the EHR were error-free, some contained hallucinations and omitted relevant information. Note, however, that the shared task does not allow data to be sent to third parties via an API.

\paragraph{Problem List Summarization (ProbSum).}
ProbSum~\cite{DBLP:conf/coling/GaoDMXCA22} is a task aimed at generating a list of problems in a patient's daily care plan based on hospital records. In the BioNLP 2023 shared task~\cite{DBLP:conf/bionlp/GaoD0A23} focused on ProbSum, the ensemble of ClinicalT5 models demonstrated robust performance~\cite{DBLP:conf/bionlp/ManakulFLRRG23}, and the approach combining Flan-T5~\cite{DBLP:journals/corr/abs-2210-11416} with GPT2XL~\cite{radford2019language} also yielded strong results~\cite{DBLP:conf/bionlp/LiWSBNKZB0N23}. 
In the experiments using the shared task dataset, LLMs adapted to the medical domain 
demonstrated performance equal to or better than medical experts~\cite{van2024adapted}.

\section{Task overview}

\begin{table*}[t!]
\tiny
\centering
\begin{tabular}{p{0.49\textwidth}|p{0.49\textwidth}}
\toprule
{\color{blue}Brief Hospital Course} & {\color{orange}Discharge Instructions}\\
\midrule
Mr. \_\_\_ is a \_\_\_ yo M with medical history significant for {\color{red}\textbackslash\textbackslash}\newline{}stage IIIb supraclavicular melanoma and prostate cancer admitted {\color{red}\textbackslash\textbackslash}\newline{}to the Acute Care Surgery Service on \_\_\_ with worsening {\color{red}\textbackslash\textbackslash}\newline{}abdominal pain, frequent stools, and subjective fevers. He was {\color{red}\textbackslash\textbackslash}\newline{}transferred from \_\_\_ for further management with a CT {\color{red}\textbackslash\textbackslash}\newline{}abdomen showing a 5 x 6 x 7 cm right mid abdominal inflammatory {\color{red}\textbackslash\textbackslash}\newline{}phlegmon. He was admitted to the surgical floor for IV {\color{red}\textbackslash\textbackslash}\newline{}antibitoics and further evaluation.{\color{red}\textbackslash\textbackslash}\newline{}{\color{red}\textbackslash\textbackslash}\newline{}Gastroenterology was consulted for duodenal thickening. Given {\color{red}\textbackslash\textbackslash}\newline{}his current infection the wall thickening is likely secondary to {\color{red}\textbackslash\textbackslash}\newline{}the infection. Repeat imaging was recommended to evaluate {\color{red}\textbackslash\textbackslash}\newline{}evolution of the phlegmon as well as outpatient colonoscopy once {\color{red}\textbackslash\textbackslash}\newline{}antibiotic treatment is complete. {\color{red}\textbackslash\textbackslash}\newline{}{\color{red}\textbackslash\textbackslash}\newline{}The remainder of the hospital course is summarized below:{\color{red}\textbackslash\textbackslash}\newline{}Neuro: The patient was alert and oriented throughout {\color{red}\textbackslash\textbackslash}\newline{}hospitalization; pain was initially managed with a IV dilaudid. {\color{red}\textbackslash\textbackslash}\newline{}He had left ankle pain and swelling consistent with gout that {\color{red}\textbackslash\textbackslash}\newline{}was managed with PO indomethacin.. {\color{red}\textbackslash\textbackslash}\newline{}CV: The patient remained stable from a cardiovascular {\color{red}\textbackslash\textbackslash}\newline{}standpoint; vital signs were routinely monitored.{\color{red}\textbackslash\textbackslash}\newline{}Pulmonary: The patient remained stable from a pulmonary {\color{red}\textbackslash\textbackslash}\newline{}standpoint. Good pulmonary toilet, early ambulation and {\color{red}\textbackslash\textbackslash}\newline{}incentive spirometry were encouraged throughout hospitalization. {\color{red}\textbackslash\textbackslash}\newline{}{\color{red}\textbackslash\textbackslash}\newline{}GI/GU/FEN: The patient was initially kept NPO.  On HD3 he was {\color{red}\textbackslash\textbackslash}\newline{}given a clear liquid diet. On HD4 he was advanced to regular {\color{red}\textbackslash\textbackslash}\newline{}diet with good tolerability. Patient's intake and output were {\color{red}\textbackslash\textbackslash}\newline{}closely monitored{\color{red}\textbackslash\textbackslash}\newline{}ID: The patient's fever curves were closely watched for signs of {\color{red}\textbackslash\textbackslash}\newline{}infection, of which there were none. He was initially given IV {\color{red}\textbackslash\textbackslash}\newline{}zosyn and transitioned to oral flagyl and ciprofloxacin upon {\color{red}\textbackslash\textbackslash}\newline{}discharge to complete a 2 week course of antibiotics. {\color{red}\textbackslash\textbackslash}\newline{}HEME: The patient's blood counts were closely watched for signs {\color{red}\textbackslash\textbackslash}\newline{}of bleeding, of which there were none.{\color{red}\textbackslash\textbackslash}\newline{}Prophylaxis: The patient received subcutaneous heparin and \_\_\_ {\color{red}\textbackslash\textbackslash}\newline{}dyne boots were used during this stay and was encouraged to get {\color{red}\textbackslash\textbackslash}\newline{}up and ambulate as early as possible.{\color{red}\textbackslash\textbackslash}\newline{}{\color{red}\textbackslash\textbackslash}\newline{}At the time of discharge, the patient was doing well, afebrile {\color{red}\textbackslash\textbackslash}\newline{}and hemodynamically stable.  The patient was tolerating a diet, {\color{red}\textbackslash\textbackslash}\newline{}ambulating, voiding without assistance, and pain was well {\color{red}\textbackslash\textbackslash}\newline{}controlled.  The patient received discharge teaching and {\color{red}\textbackslash\textbackslash}\newline{}follow-up instructions with understanding verbalized and {\color{red}\textbackslash\textbackslash}\newline{}agreement with the discharge plan. He was instructed to follow {\color{red}\textbackslash\textbackslash}\newline{}up with a colonoscopy outpatient in \_\_\_. 
&
Dr. \_\_\_,{\color{red}\textbackslash\textbackslash}\newline{}{\color{red}\textbackslash\textbackslash}\newline{}You were admitted to the Acute Care Surgery Service on \_\_\_ {\color{red}\textbackslash\textbackslash}\newline{}with abdominal pain. You had a CT scan of your abdomen that {\color{red}\textbackslash\textbackslash}\newline{}showed likely a perforated appendicitis. You were given IV {\color{red}\textbackslash\textbackslash}\newline{}antibiotics and had improvement in your symptoms. An attempt was {\color{red}\textbackslash\textbackslash}\newline{}made to drain the infection but it is not amenable to a drain at {\color{red}\textbackslash\textbackslash}\newline{}this time. You were transitioned to oral antibiotics with {\color{red}\textbackslash\textbackslash}\newline{}continued good effect.{\color{red}\textbackslash\textbackslash}\newline{}{\color{red}\textbackslash\textbackslash}\newline{}While in the hospital you had a flair up of gout in your left {\color{red}\textbackslash\textbackslash}\newline{}ankle. You were given indomethacin with improvement in your {\color{red}\textbackslash\textbackslash}\newline{}symptoms.{\color{red}\textbackslash\textbackslash}\newline{}{\color{red}\textbackslash\textbackslash}\newline{}You are now doing better, tolerating a regular diet, and ready {\color{red}\textbackslash\textbackslash}\newline{}to be discharged to home to continue your recovery.{\color{red}\textbackslash\textbackslash}\newline{}{\color{red}\textbackslash\textbackslash}\newline{}Please note the following discharge instructions:{\color{red}\textbackslash\textbackslash}\newline{}{\color{red}\textbackslash\textbackslash}\newline{}Please call your doctor or nurse practitioner or return to the {\color{red}\textbackslash\textbackslash}\newline{}Emergency Department for any of the following:{\color{red}\textbackslash\textbackslash}\newline{}*You experience new chest pain, pressure, squeezing or {\color{red}\textbackslash\textbackslash}\newline{}tightness.{\color{red}\textbackslash\textbackslash}\newline{}*New or worsening cough, shortness of breath, or wheeze.{\color{red}\textbackslash\textbackslash}\newline{}*If you are vomiting and cannot keep down fluids or your {\color{red}\textbackslash\textbackslash}\newline{}medications.{\color{red}\textbackslash\textbackslash}\newline{}*You are getting dehydrated due to continued vomiting, diarrhea, {\color{red}\textbackslash\textbackslash}\newline{}or other reasons. Signs of dehydration include dry mouth, rapid {\color{red}\textbackslash\textbackslash}\newline{}heartbeat, or feeling dizzy or faint when standing.{\color{red}\textbackslash\textbackslash}\newline{}*You see blood or dark/black material when you vomit or have a {\color{red}\textbackslash\textbackslash}\newline{}bowel movement.{\color{red}\textbackslash\textbackslash}\newline{}*You experience burning when you urinate, have blood in your {\color{red}\textbackslash\textbackslash}\newline{}urine, or experience a discharge.{\color{red}\textbackslash\textbackslash}\newline{}*Your pain in not improving within \_\_\_ hours or is not gone {\color{red}\textbackslash\textbackslash}\newline{}within 24 hours. Call or return immediately if your pain is {\color{red}\textbackslash\textbackslash}\newline{}getting worse or changes location or moving to your chest or {\color{red}\textbackslash\textbackslash}\newline{}back.{\color{red}\textbackslash\textbackslash}\newline{}*You have shaking chills, or fever greater than 101.5 degrees {\color{red}\textbackslash\textbackslash}\newline{}Fahrenheit or 38 degrees Celsius.{\color{red}\textbackslash\textbackslash}\newline{}*Any change in your symptoms, or any new symptoms that concern {\color{red}\textbackslash\textbackslash}\newline{}you.{\color{red}\textbackslash\textbackslash}\newline{}{\color{red}\textbackslash\textbackslash}\newline{}Please resume all regular home medications, unless specifically {\color{red}\textbackslash\textbackslash}\newline{}advised not to take a particular medication. Also, please take {\color{red}\textbackslash\textbackslash}\newline{}any new medications as prescribed.{\color{red}\textbackslash\textbackslash}\newline{}{\color{red}\textbackslash\textbackslash}\newline{}Please get plenty of rest, continue to ambulate several times {\color{red}\textbackslash\textbackslash}\newline{}per day, and drink adequate amounts of fluids.
\\
\bottomrule
\end{tabular}
\caption{An example of the ``{\color{blue}Brief Hospital Course}'' and ``{\color{orange}Discharge Instructions}'' sections. ``{\color{red}\textbackslash\textbackslash}'' means line breaks.}
\label{tab:brie-hospital-course-and-discharge-instructions}
\end{table*}

\begin{table*}[t!]
\tiny
\centering
\begin{tabular}{llcc}
\toprule
Section  &  Prompt & {\color{blue}Brief Hospital Course} & {\color{orange}Discharge Instructions}\\
\midrule
Name & The patient's name is provided as follows: & 1 & 1 \\
Sex & Gender details are as follows: & 2 & 2 \\
Service & The service details are as follows: & 9 & 9\\
Allergies & Information on any allergies is detailed as follows: & 7 & 6\\
Chief Complaint &  The primary reason for the visit is summarized as follows: & 3 & 3 \\
Major Surgical or Invasive Procedure & Details on any major surgeries or invasive procedures are as follows: & 8 & 7\\
History of Present Illness & An overview of the current illness's history is provided as follows: & 4 & 4 \\
Past Medical History & A summary of the patient's past medical history is as follows:& 6 & 5\\
\midrule
Pertinent Results & Clinically significant findings impacting the treatment and diagnosis are as follows: & 5 & --\\
\midrule
Medications on Admission  & Medications upon admission are detailed as follows: & -- & 8 \\
Discharge Diagnosis & The final diagnosis at discharge is as follows: & -- & 10 \\
Discharge Disposition & The disposition at discharge is provided as follows: & --& 11 \\
Discharge Condition & The patient's condition upon discharge is described as follows: & -- & 12 \\
Discharge Medications & Medications prescribed at discharge are as follows: & -- & 13 \\
\bottomrule
\end{tabular}
\caption{
Prompts for each section and their priorities in each target discharge summary.
The priority is used to order the sections in the input text.
}
\label{tab:section-prompt}
\end{table*}

\subsection{Task description}
Participants use an EHR dataset from MIMIC-IV~\cite{johnson2023mimiciv} and develop a pipeline to generate two discharge summaries: the ``{\color{blue}Brief Hospital Course}'' section for patients and the ``{\color{orange}Discharge Instructions}'' section for clinicians. Table~\ref{tab:brie-hospital-course-and-discharge-instructions} shows an example of both sections. 

\subsection{Dataset description}\label{sec:dataset}
The original datasets~\cite{xu2024discharge} include training, validation, phase I test, and phase II test sets. Participants use the training and validation sets to develop their pipeline, with the final evaluation performed on a subset of 250 samples from the phase II test set. See Appendix~\ref{app:details-of-datasets} for more details.

Note that although the datasets include metadata such as radiology reports in addition to the EHR and discharge summaries, we did not use this information in designing a simple pipeline.
For more details, see the task website\footnote{\url{https://stanford-aimi.github.io/discharge-me/}}.

We created a new split with a 4:1 training-to-validation ratio using the original training and validation sets.
Note that the EHR in the dataset contains the target texts: the ``{\color{blue}Brief Hospital Course}'' and the ``{\color{orange}Discharge Instructions}'' sections. As shown in Fig.~\ref{fig:ehr}, the ``{\color{blue}Brief Hospital Course}'' section is usually located in the middle of the discharge summary, while the ``{\color{orange}Discharge Instructions}'' section is generally located at the end of the EHR.

\begin{table*}[t!]
\scriptsize
\centering
\begin{tabular}{lllllllllll}
\toprule
Rank & Team & Overall & BLEU & ROUGE-1 & ROUGE-2 & ROUGE-L & BERTScore & Meteor & AlignScore & MEDCON \\
\midrule
1 & WisPerMed & \textbf{0.332} & \textbf{0.124} & \textbf{0.453} & \textbf{0.201} & \textbf{0.308} & \textbf{0.438} & \textbf{0.403} & \textbf{0.315} & \textbf{0.411} \\
2 & HarmonAI Lab at Yale & \underline{0.300} & 0.106$^*$ & \underline{0.423} & 0.180$^*$ & \underline{0.284} & \underline{0.412} & \underline{0.381} & 0.265$^*$ & 0.353$^*$ \\
3 & aehrc & 0.297$^*$ & 0.097 & \underline{0.414} & \underline{0.192} & \underline{0.284} & 0.383$^*$ & \underline{0.398} & 0.274$^*$ & 0.332$^*$ \\
4 & EPFL-MAKE & 0.289$^*$ & 0.098 & \underline{0.444} & 0.155 & 0.262$^*$ & \underline{0.399} & 0.336$^*$ & 0.255$^*$ & 0.360$^*$ \\
5 & UF-HOBI & 0.286$^*$ & 0.102$^*$ & 0.401$^*$ & 0.174$^*$ & 0.275$^*$ & \underline{0.395} & 0.289 & \underline{0.296} & 0.355$^*$ \\
6 & de ehren & 0.284$^*$ & 0.097 & 0.404$^*$ & 0.166$^*$ & 0.265$^*$ & 0.389$^*$ & \underline{0.376} & 0.231 & 0.339$^*$ \\
7 & DCT\_PI & 0.277$^*$ & 0.092 & 0.401$^*$ & 0.158 & 0.256$^*$ & 0.378$^*$ & \underline{0.363} & 0.247 & 0.320 \\
8 & IgnitionInnovators & 0.253 & 0.068 & 0.370$^*$ & 0.131 & 0.245 & 0.360$^*$ & 0.314 & 0.215 & 0.324 \\
9 & {\color{red}Shimo Lab (Ours)} & 0.248 & 0.063 & 0.394$^*$ & 0.131 & 0.252$^*$ & 0.351$^*$ & 0.312 & 0.210 & 0.276 \\
10 & qub-cirdan & 0.221 & 0.024 & 0.377$^*$ & 0.106 & 0.205 & 0.300 & 0.332$^*$ & 0.174 & 0.254 \\
11 & Roux-lette & 0.206 & 0.030 & 0.319 & 0.084 & 0.182 & 0.289 & 0.287 & 0.195 & 0.265 \\
12 & UoG Siephers & 0.191 & 0.017 & 0.341 & 0.109 & 0.209 & 0.268 & 0.247 & 0.143 & 0.193 \\
13 & mike-team & 0.188 & 0.022 & 0.290 & 0.076 & 0.163 & 0.258 & 0.294 & 0.182 & 0.223 \\
14 & Ixa-UPV & 0.183 & 0.016 & 0.259 & 0.057 & 0.144 & 0.282 & 0.284 & 0.210 & 0.215 \\
15 & MLBMIKABR & 0.170 & 0.039 & 0.210 & 0.092 & 0.131 & 0.186 & 0.306 & 0.205 & 0.191 \\
16 & cyq & 0.104 & 0.002 & 0.197 & 0.016 & 0.106 & 0.179 & 0.106 & 0.132 & 0.091 \\
17 & AIMI-Baseline & 0.102 & 0.015 & 0.126 & 0.052 & 0.113 & 0.138 & 0.089 & 0.167 & 0.121 \\
\bottomrule
\end{tabular}
\caption{
The evaluation metrics values for the final test data. 
The higher values are better, and the highest value is highlighted in \textbf{bold}. 
Values that are at least $90\%$ of the highest value are \underline{underlined}, and values that are at least $80\%$ of the highest value are marked with $(\ast)$.
}
\label{tab:leader-board}
\end{table*}

\begin{table*}[t!]
\tiny
\centering
\begin{tabular}{p{0.49\textwidth}|p{0.49\textwidth}}
\toprule
{\color{blue}Brief Hospital Course} & {\color{orange}Discharge Instructions}\\
\midrule
Mr. \_\_\_ is a \_\_ yo M with PMHx significant for stage IIIb supraclavicular melanoma s/p supraclavicle and right anterior neck dissection and prostate cancer presenting with abdominal pain. The remainder of the hospital course is summarized below: - Neuro: The patient was alert and oriented throughout hospitalization; pain was initially managed with a IV dilaudid. He had left ankle pain and swelling consistent with gout that was managed with PO indomethacin. CV: The patient remained stable from a cardiovascular standpoint; vital signs were routinely monitored. Pulmonary: - The patient stayed stable from an pulmonary standpoint. Good pulmonary toilet, early ambulation and incentive spirometry were encouraged throughout hospitalization - GI/GU/FEN: The patient is initially kept NPO. On HD3 he was given a clear liquid diet. On HD4 he was advanced to regular diet with good tolerability. - Patient's intake and output were closely monitored ID: The patient's fever curves were closely watched for signs of infection, of which there were none. He was initially given IV zosyn and transitioned to oral flagyl and ciprofloxacin upon discharge to complete a 2 week course of antibiotics. HEME: The patient received subcutaneous heparin and dyne boots were used during this stay and was encouraged to get up and ambulate as early as possible. At the time of discharge, the patient was doing well, afebrile and hemodynamically stable. The patient was tolerating a diet, ambulating, voiding without assistance, and pain was well controlled. The patient received discharge teaching and follow-up instructions with understanding verbalized and agreement with the discharge plan. He was instructed to follow up with a colonoscopy outpatient in \_\_\_\_\_\_\_\_\_\_\_\_\_\_\_\_\_\_\_\_\_\_\_\_\_\_\_\_\_\_\_
&
Dear Mr. \_\_\_, You were admitted to the hospital with abdominal pain. You were found to have a perforated appendicitis. You were treated with bowel rest and intravenous antibiotics. You are now ready to be discharged home to continue your recovery with the following instructions: Please call your doctor or nurse practitioner or return to the Emergency Department for any of the following: *You experience new chest pain, pressure, squeezing or tightness. *New or worsening cough, shortness of breath, or wheeze. *If you are vomiting and cannot keep down fluids or your medications. *You are getting dehydrated due to continued vomiting, diarrhea, or other reasons. Signs of dehydration include dry mouth, rapid heartbeat, or feeling dizzy or faint when standing. *You see blood or dark/black material when you vomit or have a bowel movement. *You experience burning when you urinate, have blood in your urine, or experience a discharge. *Your pain in not improving within 12 hours or is not gone within 24 hours. Call or return immediately if your pain is getting worse or changes location or moving to your chest or back. *You have shaking chills, or fever greater than 101.5 degrees Fahrenheit or 38 degrees Celsius. *Any change in your symptoms, or any new symptoms that concern you. Please resume all regular home medications, unless specifically advised not to take a particular medication. Also, please take any new medications as prescribed. Please get plenty of rest, continue to ambulate several times per day, and drink adequate amounts of fluids. Avoid lifting weights greater than \_\_- lbs until you follow-up with your surgeon. Avoid driving or operating heavy machinery while taking pain medications. 
\\
\bottomrule
\end{tabular}
\caption{Our generated texts for the ``{\color{blue}Brief Hospital Course}'' and ``{\color{orange}Discharge Instructions}'' sections in Table~\ref{tab:brie-hospital-course-and-discharge-instructions}.}
\label{tab:gen-text}
\end{table*}

\subsection{Evaluation metrics}
In this task, the following eight evaluation metrics\footnote{\url{https://github.com/Stanford-AIMI/discharge-me/tree/main/scoring}.} are used to compare the generated texts with the target texts: BLEU-4~\cite{papineni-etal-2002-bleu}, ROUGE-1, ROUGE-2, ROUGE-L~\cite{lin-2004-rouge}, BERTScore~\cite{DBLP:conf/iclr/ZhangKWWA20}, METEOR~\cite{DBLP:conf/acl/BanerjeeL05}, AlignScore~\cite{DBLP:conf/acl/ZhaYLH23}, MEDCON~\cite{Yim2023}. The overall score is calculated by first averaging the scores for each target, and then averaging these values.

\section{Pipeline}
\subsection{Input text preparation}\label{sec:input-text}
We removed the target discharge summaries from the EHR as preprocessing.
As shown in Fig.~\ref{fig:ehr}, the EHR contains redundant line breaks and detailed data. When the EHR is used directly as input text, this redundancy can increase the length of the input text. To mitigate this, we removed the noise from the EHR and selectively extracted the relevant sections for each target, thus avoiding the excessive length of the input text\footnote{The criteria for section selection are ad hoc, as mentioned in the Limitations section.}. These sections were selected by excluding those with detailed data, such as timestamps\footnote{Although the ``Pertinent Results'' section contains timestamps, we exclude them and use this section as input for the ``{\color{blue}Brief Hospital Course}'' section. See the Appendix~\ref{app:detail-section-proc}  for details.}, or those without specific information, such as the ``Admission Date'' section. Note that, in the case of preparing the input text for the model generating the ``{\color{blue}Brief Hospital Course}'' section, given the actual workflow of writing discharge summaries, we did not use the sections following this section in the input text.

For sections extracted from the EHR, we added an explanatory prompt to the beginning of each section and then concatenated the sections using the \texttt{<sep>} tokens to create the final input text.
Table~\ref{tab:section-prompt} shows the prompts and priorities of the selected sections used in the input text for each target discharge summary.
The sections in the input text were ordered according to the specified priorities, rather than their original order in the EHR.
The input text was truncated if it exceeded the maximum text length\footnote{1596 tokens}.

In Appendix~\ref{app:input-text-details}, examples of input texts are shown in Tables~\ref{tab:biref-hospital-course-input} and~\ref{tab:discharge-instructions-input}, respectively, for ``{\color{blue}Brief Hospital Course}'' and ``{\color{orange}Discharge Instructions}''.
These input texts were prepared from the EHR in Fig.~\ref{fig:ehr}.
Histograms of the length of the input text are shown in Fig.~\ref{fig:stats-input}.

\subsection{Target text preparation}\label{sec:target-text}
As shown in Table~\ref{tab:brie-hospital-course-and-discharge-instructions}, the target texts contain many unnecessary line breaks. To prevent the line breaks from hindering the learning of the model, we removed them during preprocessing. 
In Appendix~\ref{app:clearned-target}, the texts before and after preprocessing for 
``{\color{blue}Brief Hospital Course}'' are shown in Table~\ref{tab:brie-hospital-course-cleaned-target}, and those for ``{\color{orange}Discharge Instructions}'' are shown in Table~\ref{tab:discharge-instructions-clearned-target}.
Histograms of the length of the target text are shown in Fig.~\ref{fig:stats-target}.

\subsection{Text generation}
Using the input and target texts prepared in Sections~\ref{sec:input-text} and~\ref{sec:target-text}, we performed LoRA~\cite{DBLP:conf/iclr/HuSWALWWC22} fine-tuning on the ClinicalT5-large\footnote{\url{https://huggingface.co/luqh/ClinicalT5-large}} model published by \citet{DBLP:conf/emnlp/LuDN22}. 
The ClinicalT5-large model has 770M parameters with $24$ layers.
In Appendix~\ref{app:fine-tuning},
the hyperparameters for fine-tuning and LoRA are shown in Tables~\ref{tab:params} and~\ref{tab:lora-params}.
The hyperparameters to generate each target discharge summary are shown in Table~\ref{tab:gen-params}.

\section{Experiments}
\subsection{Results for the final test data}
Table~\ref{tab:leader-board} presents the evaluation metrics values of the participating teams for the final test data.
While our method did not achieve the highest scores of \textit{WisPerMed}~\cite{DBLP:journals/corr/abs-2405-11255}, it demonstrated relatively good performance in ROUGE-1, ROUGE-L, and BERTScore. In particular, we achieved a ROUGE-1 score of $0.394$, which is comparable to top solutions such as those of \textit{HarmonAI Lab at Yale} and \textit{aehrc}. 

\subsection{Qualitative observation}
Table~\ref{tab:gen-text} presents the summaries generated by our pipeline from the EHR for the target summaries in Table~\ref{tab:brie-hospital-course-and-discharge-instructions}.
While the detailed progress reports and discharge instructions may differ, the overall gist remains the same. 
In addition, unnecessary line breaks that were present in the original target summaries do not appear in the generated summaries.

\section{Conclusion}
We presented our approach to the shared task ``Discharge Me!'' at the BioNLP Workshop 2024. 
Extracting the relevant sections from the EHR, we added explanatory prompts to these sections and concatenated them with \texttt{<sep>} tokens to create the input text. We then performed LoRA fine-tuning on the ClinicalT5-large model.
On the final test data, our approach achieved a ROUGE-1 score of $0.394$, which is comparable to the top solutions. 

\section*{Limitations}
\begin{itemize}
    \item Our pipeline cannot be applied to an EHR with different formats, resulting in a lack of generalizability. In fact, even in this shared task dataset, the lack of consistency in the original data sometimes makes it impossible to extract sections, resulting in incomplete summaries.
    \item When preparing the input text, adding prompts for each extracted section results in a longer length than simply concatenating sections with \texttt{<sep>} tokens.
    \item The effectiveness of our pipeline is not tested against other text generation models such as BioMistral-7B~\cite{DBLP:journals/corr/abs-2402-10373} and the ClinicalT5-large model published by \citet{DBLP:conf/chil/LehmanHMWSZNSJA23}.
    \item While the selection and prioritization of the EHR sections used in the input text is somewhat ad-hoc, since extensive experiments would be required to compare the selection and prioritization, we did not conduct them in this study due to time and resource constraints.
    \item  While the cleaned target texts are used for training, the original target texts with many line breaks are used for evaluation. This leads to a discrepancy between the target text distributions of training and evaluation.
\end{itemize}

\section*{Ethics Statement}
We conducted our research with careful consideration of data use and in accordance with the Data Use Agreement\footnote{\url{https://physionet.org/content/discharge-me/view-dua/1.3/}}. It is prohibited to identify individuals or organizations from the examples presented in the paper.

\section*{Acknowledgements}
We would like to thank the Program Committee for their thorough review and valuable suggestions.
This work was supported by JST SPRING, Grant Number JPMJSP2110. This study was partially supported by JST CREST JPMJCR21N3.

\bibliography{custom}
\bibliographystyle{acl_natbib}

\appendix
\begin{table}[t!]
\centering
\begin{tabular}{lccl}
\toprule
Dataset & Numbers \\
\midrule
Training & $68{,}785$\\
Validation & $14{,}719$\\
Phase I Test & $14{,}702$\\
Phase II Test & $10{,}962$\\
\midrule
Total & $109{,}168$\\
\bottomrule
\end{tabular}
\caption{The number of samples for the data splits.}
\label{tab:original-dataset-numbers}
\end{table}

\begin{table*}[t!]
\tiny
\centering
\begin{tabular}{p{0.95\textwidth}}
\toprule
{\color{green} The patient's name is provided as follows:} \_\_\_.\pink{\textbf{<sep>}}{\color{green} Gender details are as follows:} Male.\pink{\textbf{<sep>}}{\color{green} The primary reason for the visit is summarized as follows:} Abdominal pain.\pink{\textbf{<sep>}}{\color{green} An overview of the current illness's history is provided as follows:} Mr. \_\_\_ is a \_\_\_ male with PMHx significant for stage IIIb supraclavicular melanoma s/p supraclavicular and right anterior neck dissection and prostate cancer s/p radical prostatectomy, presenting with abdominal pain. The pain began two weeks ago and has been worsening in the last few days. His pain is localized to the right periumbilical region. He endorses having chills, inability to sleep and eat due to pain and a 6 lb weight loss in the past few days. He has been passing flatus and having loose and frequent non-bloody stools. He has also been having night sweats. He denies fever, nausea or vomiting. He was seen at \_\_\_ and transferred to \_\_\_ ED for further management after his CT abdomen showed a 5 x 6 x 7 cm right mid abdominal inflammatory phlegmon. His last colonoscopy was \_\_\_ years ago which revealed some polyps. He also complains of left ankle pain after a fall and has been taking ibuprofen.\pink{\textbf{<sep>}}{\color{blue} Clinically significant findings impacting the treatment and diagnosis are as follows:} MCV-89 MCH-29.4 MCHC-32.9 RDW-13.0 RDWSD-42.2 Plt \_\_\_ MCV-88 MCH-29.8 MCHC-34.1 RDW-12.7 RDWSD-40.3 Plt \_\_\_ MCV-88 MCH-29.9 MCHC-33.8 RDW-12.5 RDWSD-40.2 Plt \_\_\_ MCV-88 MCH-29.5 MCHC-33.5 RDW-12.4 RDWSD-39.8 Plt \_\_\_ K-3.8 Cl-103 HCO3-24 AnGap-18 K-3.6 Cl-99 HCO3-24 AnGap-20 K-3.5 Cl-97 HCO3-24 AnGap-19 K-3.8 Cl-96 HCO3-21* AnGap-22* Glucose-NEG Ketone-80 Bilirub-NEG Urobiln-NEG pH-6.0 Leuks-NEG Epi-0 **FINAL REPORT \_\_\_. URINE CULTURE (Final \_\_\_: NO GROWTH. RADIOLOGY: * Phlegmon/ multiloculated fluid collection with surrounding. extensive inflammatory changes is indistinguishable from the distal portion of the appendix. Findings are concerning for perforated appendicitis. Possibility of underlying mass is difficult to exclude, particularly in this patient with history of melanoma. * Duodenal wall thickening may be inflammatory secondary to the. adjacent phlegmon. * Duodenum does not cross the midline, consistent with. intestinal malrotation. * Cholelithiasis. * Nonspecific bulbous appearance of the uncinate process of the. pancreas without discrete lesion identified. No pancreatic ductal dilatation. * Unorganized fluid/phlegmonous collection within the right. lower quadrant, surrounding the appendix, appears minimally enlarged since the reference study from \_\_\_. The findings favor ruptured appendicitis or a ruptured appendiceal mucocele. A neoplastic source relating to known history of melanoma would be atypical. Continued short-term imaging surveillance is recommended. * Congenital bowel malrotation, without volvulus or. obstruction. * Cholelithiasis. The remainder of the hospital course is summarized below: Neuro: The patient was alert and oriented throughout hospitalization; pain was initially managed with a IV dilaudid. He had left ankle pain and swelling consistent with gout that was managed with PO indomethacin.. CV: The patient remained stable from a cardiovascular standpoint; vital signs were routinely monitored. Pulmonary: The patient remained stable from a pulmonary standpoint. Good pulmonary toilet, early ambulation and incentive spirometry were encouraged throughout hospitalization. GI/GU/FEN: The patient was initially kept NPO. On HD3 he was given a clear liquid diet. On HD4 he was advanced to regular diet with good tolerability. Patient's intake and output were closely monitored. ID: The patient's fever curves were closely watched for signs of infection, of which there were none. He was initially given IV zosyn and transitioned to oral flagyl and ciprofloxacin upon discharge to complete a 2 week course of antibiotics. HEME: The patient's blood counts were closely watched for signs of bleeding, of which there were none. Prophylaxis: The patient received subcutaneous heparin and \_\_\_ dyne boots were used during this stay and was encouraged to get up and ambulate as early as possible. At the time of discharge, the patient was doing well, afebrile and hemodynamically stable. The patient was tolerating a diet, ambulating, voiding without assistance, and pain was well controlled. The patient received discharge teaching and follow-up instructions with understanding verbalized and agreement with the discharge plan. He was instructed to follow up with a colonoscopy outpatient in \_\_\_.\pink{\textbf{<sep>}}{\color{green} A summary of the patient's past medical history is as follows:} 1.right acoustic neuroma (deafness right ear) 2.s/p repair right biceps tendon rupture (\_\_\_) 3.s/p right supraclavicular lymph node biopsy (\_\_\_). PAST MEDICAL HISTORY: Stage IIIb melanoma diagnosed in \_\_\_ with findings of a positive right supraclavicular node, status post right anterior neck dissection revealing \_\_\_ additional positive nodes. He had adjuvant interferon therapy with Dr. \_\_\_ completed in \_\_\_, 36 weeks of this treatment. Bicep tendon repair. Acoustic neuroma followed with serial MRIs.\pink{\textbf{<sep>}}{\color{green} Information on any allergies is detailed as follows:} Codeine / Levaquin.\pink{\textbf{<sep>}}{\color{green} Details on any major surgeries or invasive procedures are as follows:} None.\pink{\textbf{<sep>}}{\color{green} The service details are provided as follows:} SURGERY.\\
\bottomrule
\end{tabular}
\caption{
Input text from the EHR shown in Fig.~\ref{fig:ehr} to generate the ``{\color{blue}Brief Hospital Course}'' section. The prompts used in both targets are highlighted in {\color{green}green} and the prompt used only for ``{\color{blue}Brief Hospital Course}'' is highlighted in {\color{blue}blue}.
}
\label{tab:biref-hospital-course-input}
\end{table*}

\begin{table*}[t!]
\tiny
\centering
\begin{tabular}{p{0.95\textwidth}}
\toprule
{\color{green} The patient's name is provided as follows:} \_\_\_.\pink{\textbf{<sep>}}{\color{green} Gender details are as follows:} Male.\pink{\textbf{<sep>}}{\color{green} The primary reason for the visit is summarized as follows:} Abdominal pain.\pink{\textbf{<sep>}}{\color{green} An overview of the current illness's history is provided as follows:} Mr. \_\_\_ is a \_\_\_ male with PMHx significant for stage IIIb supraclavicular melanoma s/p supraclavicular and right anterior neck dissection and prostate cancer s/p radical prostatectomy, presenting with abdominal pain. The pain began two weeks ago and has been worsening in the last few days. His pain is localized to the right periumbilical region. He endorses having chills, inability to sleep and eat due to pain and a 6 lb weight loss in the past few days. He has been passing flatus and having loose and frequent non-bloody stools. He has also been having night sweats. He denies fever, nausea or vomiting. He was seen at \_\_\_ and transferred to \_\_\_ ED for further management after his CT abdomen showed a 5 x 6 x 7 cm right mid abdominal inflammatory phlegmon. His last colonoscopy was \_\_\_ years ago which revealed some polyps. He also complains of left ankle pain after a fall and has been taking ibuprofen.\pink{\textbf{<sep>}}{\color{green} A summary of the patient's past medical history is as follows:} 1.right acoustic neuroma (deafness right ear) 2.s/p repair right biceps tendon rupture (\_\_\_) 3.s/p right supraclavicular lymph node biopsy (\_\_\_). PAST MEDICAL HISTORY: Stage IIIb melanoma diagnosed in \_\_\_ with findings of a positive right supraclavicular node, status post right anterior neck dissection revealing \_\_\_ additional positive nodes. He had adjuvant interferon therapy with Dr. \_\_\_ completed in \_\_\_, 36 weeks of this treatment. Bicep tendon repair. Acoustic neuroma followed with serial MRIs.\pink{\textbf{<sep>}}{\color{green} Information on any allergies is detailed as follows:} Codeine / Levaquin.\pink{\textbf{<sep>}}{\color{green} Details on any major surgeries or invasive procedures are as follows:} None.\pink{\textbf{<sep>}}{\color{orange} Medications upon admission are detailed as follows:} tadalafil (CIALIS) 5 mg daily PRN indomethacin 25 mg capsule TID.\pink{\textbf{<sep>}}{\color{green} The service details are provided as follows:} SURGERY.\pink{\textbf{<sep>}}{\color{orange} The final diagnosis at discharge is as follows:} Perforated appendicitis.\pink{\textbf{<sep>}}{\color{orange} The disposition at discharge is provided as follows:} Home.\pink{\textbf{<sep>}}{\color{orange} The patient's condition upon discharge is described as follows:} Mental Status is Clear and coherent. Level of Consciousness is Alert and interactive. Activity Status is Ambulatory - Independent.\pink{\textbf{<sep>}}{\color{orange} Medications prescribed at discharge are as follows:} * Acetaminophen 1000 mg PO TID. Do not exceed 4 grams/ 24 hours. * Ciprofloxacin HCl 500 mg PO Q12H. monitor for s/sx of allergic reaction RX *ciprofloxacin HCl 500 mg 1 tablet(s) by mouth twice a day. Disp \#*20 Tablet Refills:*0 * Indomethacin 25 mg PO TID. RX *indomethacin 25 mg 1 capsule(s) by mouth three times a day. Disp \#*42 Capsule Refills:*0 * MetroNIDAZOLE 500 mg PO Q8H. RX *metronidazole 500 mg 1 tablet(s) by mouth three times a day. Disp \#*30 Tablet Refills:*0.\\
\bottomrule
\end{tabular}
\caption{
Input text from the EHR in Fig.~\ref{fig:ehr} to generate the ``{\color{orange}Discharge Instructions}'' section. The prompts used in both targets are highlighted in {\color{green}green} and the prompts used only for ``{\color{orange}Discharge Instructions}'' are highlighted in {\color{orange}orange}.
}
\label{tab:discharge-instructions-input}
\end{table*}

\section{Details of datasets}\label{app:details-of-datasets}
In this task, we use the dataset created by the MIMIC-IV's submodules MIMIC-IV-ED~\cite{johnson2023mimicived} and MIMIC-IV-Note~\cite{johnson2023mimicivnote}. The dataset is available on PhysioNet~\cite{Goldberger2000}, and its use requires completion of the CITI\footnote{\url{https://about.citiprogram.org/}} training and credentialing process. Table~\ref{tab:original-dataset-numbers} lists the number of samples for the data splits.

\section{Details of input text}\label{app:input-text-details}
\begin{figure*}[!t]
    \centering
        \begin{minipage}{0.5\linewidth}
        \vspace*{5mm}
        \centering
        \includegraphics[width=\linewidth]{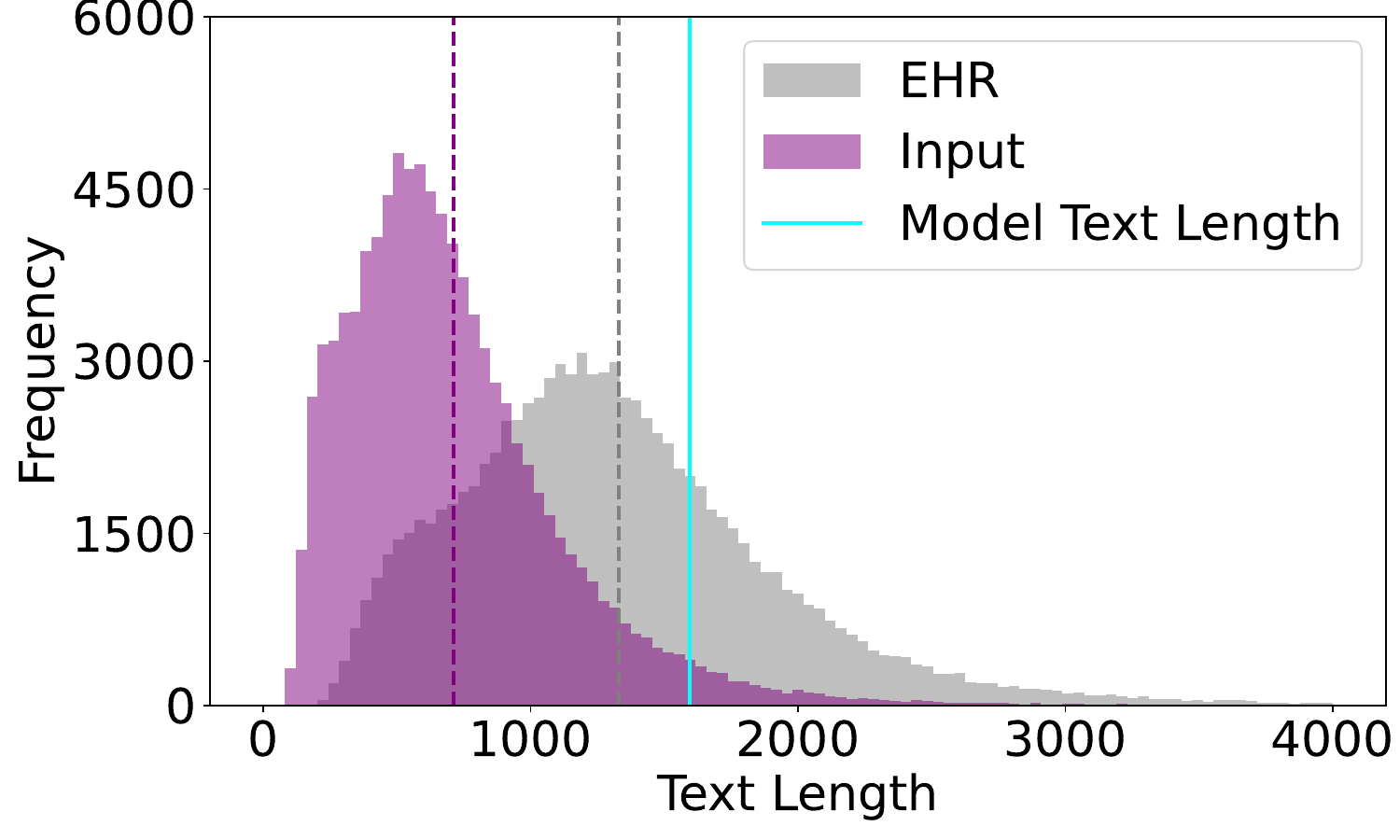}
        \subcaption{{\color{blue}Brief Hospital Course}}
        \label{fig:input-length-brief}
    \end{minipage}\hfill
    \begin{minipage}{0.5\linewidth}
        \vspace*{5mm}
        \centering
        \includegraphics[width=\linewidth]{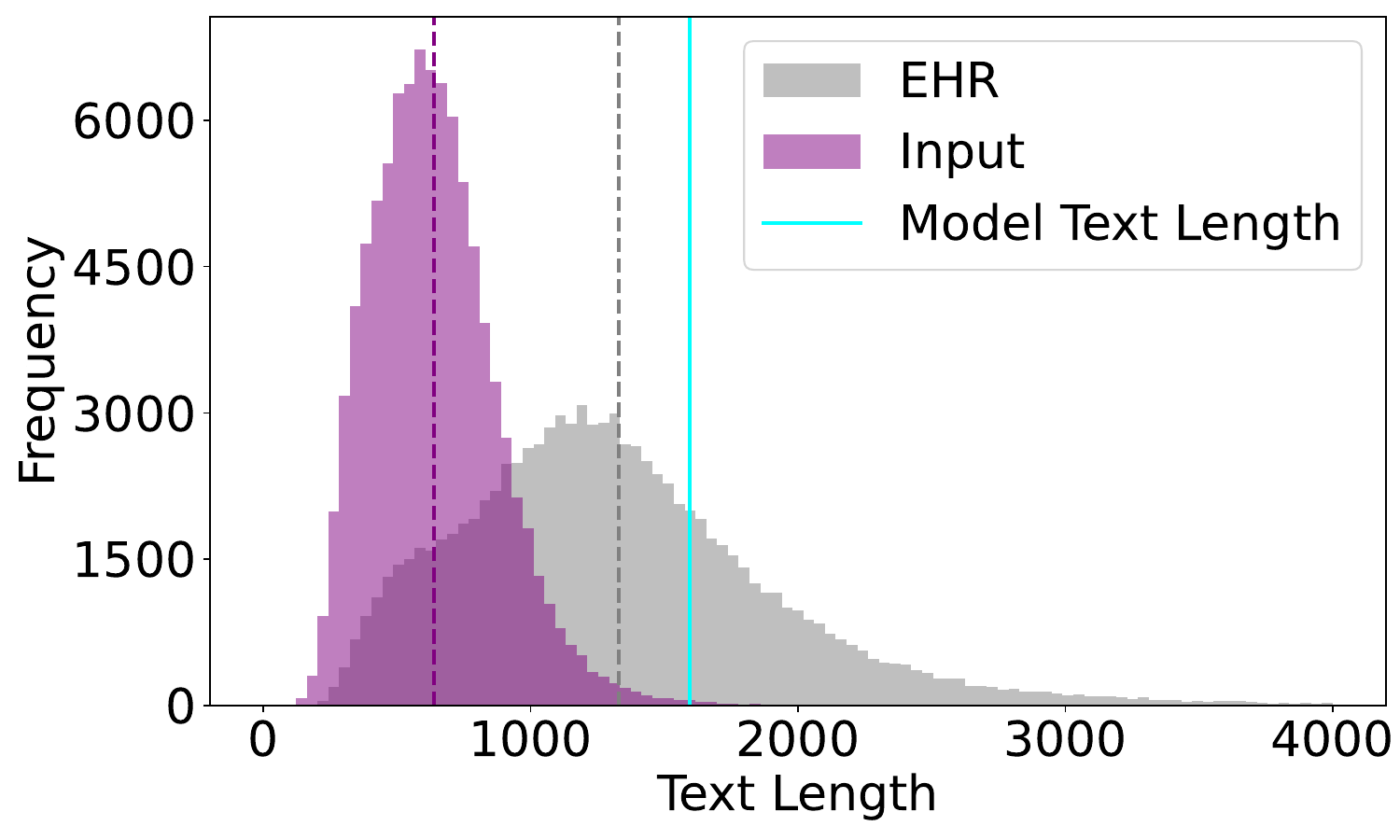}
        \subcaption{{\color{orange}Discharge Instructions}}
        \label{fig:input-lenght-instructs}
    \end{minipage}
    \caption{
Histograms of the text length (in tokens) of the EHR and the input texts for the training and validation sets. The dashed line is the mean. The maximum text length is 1596 tokens, and see Table~\ref{tab:model-text-length} in Appendix~\ref{app:fine-tuning} for more details.
}
    \label{fig:stats-input}
\end{figure*}

\begin{table*}[t!]
\centering
\begin{tabular}{lcccc}
\toprule
& \multicolumn{2}{c}{{\color{blue}Brief Hospital Course}} & \multicolumn{2}{c}{{\color{orange}Discharge instructions}}\\
& EHR & Input & EHR & Input \\
\midrule
Min & 101 & 80 & 101 & 115 \\
Max & 8725 & 5664 & 8725 & 5774\\
Mean & 1330 & 554 & 1330 & 639 \\
\bottomrule
\end{tabular}
\caption{Statistical information (in tokens) for histograms in Fig.~\ref{fig:stats-input}.}
\label{tab:stats-input}
\end{table*}
This section first explains the detailed preprocessing required to create input text from the EHR. It then provides examples and statistical information before and after preprocessing.

\begin{table*}[t!]
\tiny
\centering
\begin{tabular}{p{0.49\textwidth}|p{0.49\textwidth}}
\toprule
Text & Cleaned Text \\
\midrule
Mr. \_\_\_ is a \_\_\_ yo M with medical history significant for {\color{red}\textbackslash\textbackslash}\newline{}stage IIIb supraclavicular melanoma and prostate cancer admitted {\color{red}\textbackslash\textbackslash}\newline{}to the Acute Care Surgery Service on \_\_\_ with worsening {\color{red}\textbackslash\textbackslash}\newline{}abdominal pain, frequent stools, and subjective fevers. He was {\color{red}\textbackslash\textbackslash}\newline{}transferred from \_\_\_ for further management with a CT {\color{red}\textbackslash\textbackslash}\newline{}abdomen showing a 5 x 6 x 7 cm right mid abdominal inflammatory {\color{red}\textbackslash\textbackslash}\newline{}phlegmon. He was admitted to the surgical floor for IV {\color{red}\textbackslash\textbackslash}\newline{}antibitoics and further evaluation.{\color{red}\textbackslash\textbackslash}\newline{}{\color{red}\textbackslash\textbackslash}\newline{}Gastroenterology was consulted for duodenal thickening. Given {\color{red}\textbackslash\textbackslash}\newline{}his current infection the wall thickening is likely secondary to {\color{red}\textbackslash\textbackslash}\newline{}the infection. Repeat imaging was recommended to evaluate {\color{red}\textbackslash\textbackslash}\newline{}evolution of the phlegmon as well as outpatient colonoscopy once {\color{red}\textbackslash\textbackslash}\newline{}antibiotic treatment is complete. {\color{red}\textbackslash\textbackslash}\newline{}{\color{red}\textbackslash\textbackslash}\newline{}The remainder of the hospital course is summarized below:{\color{red}\textbackslash\textbackslash}\newline{}Neuro: The patient was alert and oriented throughout {\color{red}\textbackslash\textbackslash}\newline{}hospitalization; pain was initially managed with a IV dilaudid. {\color{red}\textbackslash\textbackslash}\newline{}He had left ankle pain and swelling consistent with gout that {\color{red}\textbackslash\textbackslash}\newline{}was managed with PO indomethacin.. {\color{red}\textbackslash\textbackslash}\newline{}CV: The patient remained stable from a cardiovascular {\color{red}\textbackslash\textbackslash}\newline{}standpoint; vital signs were routinely monitored.{\color{red}\textbackslash\textbackslash}\newline{}Pulmonary: The patient remained stable from a pulmonary {\color{red}\textbackslash\textbackslash}\newline{}standpoint. Good pulmonary toilet, early ambulation and {\color{red}\textbackslash\textbackslash}\newline{}incentive spirometry were encouraged throughout hospitalization. {\color{red}\textbackslash\textbackslash}\newline{}{\color{red}\textbackslash\textbackslash}\newline{}GI/GU/FEN: The patient was initially kept NPO.  On HD3 he was {\color{red}\textbackslash\textbackslash}\newline{}given a clear liquid diet. On HD4 he was advanced to regular {\color{red}\textbackslash\textbackslash}\newline{}diet with good tolerability. Patient's intake and output were {\color{red}\textbackslash\textbackslash}\newline{}closely monitored{\color{red}\textbackslash\textbackslash}\newline{}ID: The patient's fever curves were closely watched for signs of {\color{red}\textbackslash\textbackslash}\newline{}infection, of which there were none. He was initially given IV {\color{red}\textbackslash\textbackslash}\newline{}zosyn and transitioned to oral flagyl and ciprofloxacin upon {\color{red}\textbackslash\textbackslash}\newline{}discharge to complete a 2 week course of antibiotics. {\color{red}\textbackslash\textbackslash}\newline{}HEME: The patient's blood counts were closely watched for signs {\color{red}\textbackslash\textbackslash}\newline{}of bleeding, of which there were none.{\color{red}\textbackslash\textbackslash}\newline{}Prophylaxis: The patient received subcutaneous heparin and \_\_\_ {\color{red}\textbackslash\textbackslash}\newline{}dyne boots were used during this stay and was encouraged to get {\color{red}\textbackslash\textbackslash}\newline{}up and ambulate as early as possible.{\color{red}\textbackslash\textbackslash}\newline{}{\color{red}\textbackslash\textbackslash}\newline{}At the time of discharge, the patient was doing well, afebrile {\color{red}\textbackslash\textbackslash}\newline{}and hemodynamically stable.  The patient was tolerating a diet, {\color{red}\textbackslash\textbackslash}\newline{}ambulating, voiding without assistance, and pain was well {\color{red}\textbackslash\textbackslash}\newline{}controlled.  The patient received discharge teaching and {\color{red}\textbackslash\textbackslash}\newline{}follow-up instructions with understanding verbalized and {\color{red}\textbackslash\textbackslash}\newline{}agreement with the discharge plan. He was instructed to follow {\color{red}\textbackslash\textbackslash}\newline{}up with a colonoscopy outpatient in \_\_\_.
&
Mr. \_\_\_ is a \_\_\_ yo M with medical history significant for stage IIIb supraclavicular melanoma and prostate cancer admitted to the Acute Care Surgery Service on \_\_\_ with worsening abdominal pain, frequent stools, and subjective fevers. He was transferred from \_\_\_ for further management with a CT abdomen showing a 5 x 6 x 7 cm right mid abdominal inflammatory phlegmon. He was admitted to the surgical floor for IV antibitoics and further evaluation.{\color{red}\textbackslash\textbackslash}\newline{}{\color{red}\textbackslash\textbackslash}\newline{}Gastroenterology was consulted for duodenal thickening. Given his current infection the wall thickening is likely secondary to the infection. Repeat imaging was recommended to evaluate evolution of the phlegmon as well as outpatient colonoscopy once antibiotic treatment is complete.{\color{red}\textbackslash\textbackslash}\newline{}{\color{red}\textbackslash\textbackslash}\newline{}The remainder of the hospital course is summarized below:{\color{red}\textbackslash\textbackslash}\newline{}Neuro: The patient was alert and oriented throughout hospitalization; pain was initially managed with a IV dilaudid. He had left ankle pain and swelling consistent with gout that was managed with PO indomethacin.. CV: The patient remained stable from a cardiovascular standpoint; vital signs were routinely monitored.{\color{red}\textbackslash\textbackslash}\newline{}Pulmonary: The patient remained stable from a pulmonary standpoint. Good pulmonary toilet, early ambulation and incentive spirometry were encouraged throughout hospitalization.{\color{red}\textbackslash\textbackslash}\newline{}{\color{red}\textbackslash\textbackslash}\newline{}GI/GU/FEN: The patient was initially kept NPO. On HD3 he was given a clear liquid diet. On HD4 he was advanced to regular diet with good tolerability. Patient's intake and output were closely monitored{\color{red}\textbackslash\textbackslash}\newline{}ID: The patient's fever curves were closely watched for signs of infection, of which there were none. He was initially given IV zosyn and transitioned to oral flagyl and ciprofloxacin upon discharge to complete a 2 week course of antibiotics. HEME: The patient's blood counts were closely watched for signs of bleeding, of which there were none.{\color{red}\textbackslash\textbackslash}\newline{}Prophylaxis: The patient received subcutaneous heparin and \_\_\_ dyne boots were used during this stay and was encouraged to get up and ambulate as early as possible.{\color{red}\textbackslash\textbackslash}\newline{}{\color{red}\textbackslash\textbackslash}\newline{}At the time of discharge, the patient was doing well, afebrile and hemodynamically stable. The patient was tolerating a diet, ambulating, voiding without assistance, and pain was well controlled. The patient received discharge teaching and follow-up instructions with understanding verbalized and agreement with the discharge plan. He was instructed to follow up with a colonoscopy outpatient in \_\_\_.
\\
\bottomrule
\end{tabular}
\caption{
The text of the ``{\color{blue}Brief Hospital Course}'' section in Table~\ref{tab:brie-hospital-course-and-discharge-instructions} and its cleaned text by preprocessing. ``{\color{red}\textbackslash\textbackslash}'' means line breaks.
}
\label{tab:brie-hospital-course-cleaned-target}
\end{table*}

\begin{table*}[t!]
\tiny
\centering
\begin{tabular}{p{0.49\textwidth}|p{0.49\textwidth}}
\toprule
Text & Cleaned Text \\
\midrule
Dr. \_\_\_,{\color{red}\textbackslash\textbackslash}\newline{}{\color{red}\textbackslash\textbackslash}\newline{}You were admitted to the Acute Care Surgery Service on \_\_\_ {\color{red}\textbackslash\textbackslash}\newline{}with abdominal pain. You had a CT scan of your abdomen that {\color{red}\textbackslash\textbackslash}\newline{}showed likely a perforated appendicitis. You were given IV {\color{red}\textbackslash\textbackslash}\newline{}antibiotics and had improvement in your symptoms. An attempt was {\color{red}\textbackslash\textbackslash}\newline{}made to drain the infection but it is not amenable to a drain at {\color{red}\textbackslash\textbackslash}\newline{}this time. You were transitioned to oral antibiotics with {\color{red}\textbackslash\textbackslash}\newline{}continued good effect.{\color{red}\textbackslash\textbackslash}\newline{}{\color{red}\textbackslash\textbackslash}\newline{}While in the hospital you had a flair up of gout in your left {\color{red}\textbackslash\textbackslash}\newline{}ankle. You were given indomethacin with improvement in your {\color{red}\textbackslash\textbackslash}\newline{}symptoms.{\color{red}\textbackslash\textbackslash}\newline{}{\color{red}\textbackslash\textbackslash}\newline{}You are now doing better, tolerating a regular diet, and ready {\color{red}\textbackslash\textbackslash}\newline{}to be discharged to home to continue your recovery.{\color{red}\textbackslash\textbackslash}\newline{}{\color{red}\textbackslash\textbackslash}\newline{}Please note the following discharge instructions:{\color{red}\textbackslash\textbackslash}\newline{}{\color{red}\textbackslash\textbackslash}\newline{}Please call your doctor or nurse practitioner or return to the {\color{red}\textbackslash\textbackslash}\newline{}Emergency Department for any of the following:{\color{red}\textbackslash\textbackslash}\newline{}*You experience new chest pain, pressure, squeezing or {\color{red}\textbackslash\textbackslash}\newline{}tightness.{\color{red}\textbackslash\textbackslash}\newline{}*New or worsening cough, shortness of breath, or wheeze.{\color{red}\textbackslash\textbackslash}\newline{}*If you are vomiting and cannot keep down fluids or your {\color{red}\textbackslash\textbackslash}\newline{}medications.{\color{red}\textbackslash\textbackslash}\newline{}*You are getting dehydrated due to continued vomiting, diarrhea, {\color{red}\textbackslash\textbackslash}\newline{}or other reasons. Signs of dehydration include dry mouth, rapid {\color{red}\textbackslash\textbackslash}\newline{}heartbeat, or feeling dizzy or faint when standing.{\color{red}\textbackslash\textbackslash}\newline{}*You see blood or dark/black material when you vomit or have a {\color{red}\textbackslash\textbackslash}\newline{}bowel movement.{\color{red}\textbackslash\textbackslash}\newline{}*You experience burning when you urinate, have blood in your {\color{red}\textbackslash\textbackslash}\newline{}urine, or experience a discharge.{\color{red}\textbackslash\textbackslash}\newline{}*Your pain in not improving within \_\_\_ hours or is not gone {\color{red}\textbackslash\textbackslash}\newline{}within 24 hours. Call or return immediately if your pain is {\color{red}\textbackslash\textbackslash}\newline{}getting worse or changes location or moving to your chest or {\color{red}\textbackslash\textbackslash}\newline{}back.{\color{red}\textbackslash\textbackslash}\newline{}*You have shaking chills, or fever greater than 101.5 degrees {\color{red}\textbackslash\textbackslash}\newline{}Fahrenheit or 38 degrees Celsius.{\color{red}\textbackslash\textbackslash}\newline{}*Any change in your symptoms, or any new symptoms that concern {\color{red}\textbackslash\textbackslash}\newline{}you.{\color{red}\textbackslash\textbackslash}\newline{}{\color{red}\textbackslash\textbackslash}\newline{}Please resume all regular home medications, unless specifically {\color{red}\textbackslash\textbackslash}\newline{}advised not to take a particular medication. Also, please take {\color{red}\textbackslash\textbackslash}\newline{}any new medications as prescribed.{\color{red}\textbackslash\textbackslash}\newline{}{\color{red}\textbackslash\textbackslash}\newline{}Please get plenty of rest, continue to ambulate several times {\color{red}\textbackslash\textbackslash}\newline{}per day, and drink adequate amounts of fluids.
 &
Dr. \_\_\_,{\color{red}\textbackslash\textbackslash}\newline{}{\color{red}\textbackslash\textbackslash}\newline{}You were admitted to the Acute Care Surgery Service on \_\_\_ with abdominal pain. You had a CT scan of your abdomen that showed likely a perforated appendicitis. You were given IV antibiotics and had improvement in your symptoms. An attempt was made to drain the infection but it is not amenable to a drain at this time. You were transitioned to oral antibiotics with continued good effect.{\color{red}\textbackslash\textbackslash}\newline{}{\color{red}\textbackslash\textbackslash}\newline{}While in the hospital you had a flair up of gout in your left ankle. You were given indomethacin with improvement in your symptoms.{\color{red}\textbackslash\textbackslash}\newline{}{\color{red}\textbackslash\textbackslash}\newline{}You are now doing better, tolerating a regular diet, and ready to be discharged to home to continue your recovery.{\color{red}\textbackslash\textbackslash}\newline{}{\color{red}\textbackslash\textbackslash}\newline{}Please note the following discharge instructions:{\color{red}\textbackslash\textbackslash}\newline{}{\color{red}\textbackslash\textbackslash}\newline{}Please call your doctor or nurse practitioner or return to the Emergency Department for any of the following:{\color{red}\textbackslash\textbackslash}\newline{}*You experience new chest pain, pressure, squeezing or tightness.{\color{red}\textbackslash\textbackslash}\newline{}*New or worsening cough, shortness of breath, or wheeze.{\color{red}\textbackslash\textbackslash}\newline{}*If you are vomiting and cannot keep down fluids or your medications.{\color{red}\textbackslash\textbackslash}\newline{}*You are getting dehydrated due to continued vomiting, diarrhea, or other reasons. Signs of dehydration include dry mouth, rapid heartbeat, or feeling dizzy or faint when standing.{\color{red}\textbackslash\textbackslash}\newline{}*You see blood or dark/black material when you vomit or have a bowel movement.{\color{red}\textbackslash\textbackslash}\newline{}*You experience burning when you urinate, have blood in your urine, or experience a discharge.{\color{red}\textbackslash\textbackslash}\newline{}*Your pain in not improving within \_\_\_ hours or is not gone within 24 hours. Call or return immediately if your pain is getting worse or changes location or moving to your chest or back.{\color{red}\textbackslash\textbackslash}\newline{}*You have shaking chills, or fever greater than 101.5 degrees Fahrenheit or 38 degrees Celsius.{\color{red}\textbackslash\textbackslash}\newline{}*Any change in your symptoms, or any new symptoms that concern you.{\color{red}\textbackslash\textbackslash}\newline{}{\color{red}\textbackslash\textbackslash}\newline{}Please resume all regular home medications, unless specifically advised not to take a particular medication. Also, please take any new medications as prescribed.{\color{red}\textbackslash\textbackslash}\newline{}{\color{red}\textbackslash\textbackslash}\newline{}Please get plenty of rest, continue to ambulate several times per day, and drink adequate amounts of fluids.
\\
\bottomrule
\end{tabular}
\caption{
The text of the ``{\color{orange} Discharge Instructions}'' section in Table~\ref{tab:brie-hospital-course-and-discharge-instructions} and its cleaned text by preprocessing. ``{\color{red}\textbackslash\textbackslash}'' means line breaks.
}
\label{tab:discharge-instructions-clearned-target}
\end{table*}

\subsection{Extraction of simple sections}
This section explains the process for extracting the ``Sex'', ``Service'', ``Allergies'', ``Chief Complaint'', and ``Major Surgical or Invasive Procedure'' sections.

To extract these sections, we used specific regular expressions such as \texttt{Sex: (\textbackslash w+)\textbackslash n}.

\subsection{Extraction of complex sections}
This section explains the process for extracting the ``History of Present Illness'', ``Past Medical History'', ``Pertinent Results'', ``Medications on Admission'', ``Discharge Medications'', ``Discharge Disposition'', ``Discharge Diagnosis'', and ''Discharge Condition'' sections.

We performed more detailed processing and pattern matching to efficiently extract the text of these sections.
For example, for the ``Discharge Condition'' section, we used the regular expression \texttt{Discharge Diagnosis:\textbackslash s$\ast$\textbackslash n(.$\ast$?)(?=Discharge Condition:)} and it matches the diagnosis text up to the ``Discharge Condition'' section.

\subsection{Detailed processing of each section}\label{app:detail-section-proc}
\paragraph{``Name''.}~The patient's name is given as ``\_\_\_'' and we used it directly.

\paragraph{``Sex''.} ~We converted ``M'' to ``Male'' and ``F'' to ``Female''.

\paragraph{``Pertinent Results''.} Timestamps in lines like ``\_\_ 08:00AM BLOOD \_\_'' were removed using regular expressions.
In addition, list sections are converted to ``*'' format to maintain text consistency and clarity.

\paragraph{``Medications on Admission''.} List sections are converted to ``*'' format to maintain text consistency and clarity.

\paragraph{``Discharge Condition''.} We changed a colon in the extracted text to ``is''. For example, ``Condition: Stable'' is changed to ``Condition is Stable''.

\paragraph{``Discharge Medications''.} List sections are converted to ``*'' format to maintain text consistency and clarity.

\subsection{Other processing}
We ensure textual continuity by replacing line breaks with spaces and trimming excess spaces. In cases where no matching text is found, the default response is designated as ``Unknown''.

\begin{figure*}[!t]
    \centering
        \begin{minipage}{0.5\linewidth}
        \vspace*{5mm}
        \centering
        \includegraphics[width=\linewidth]{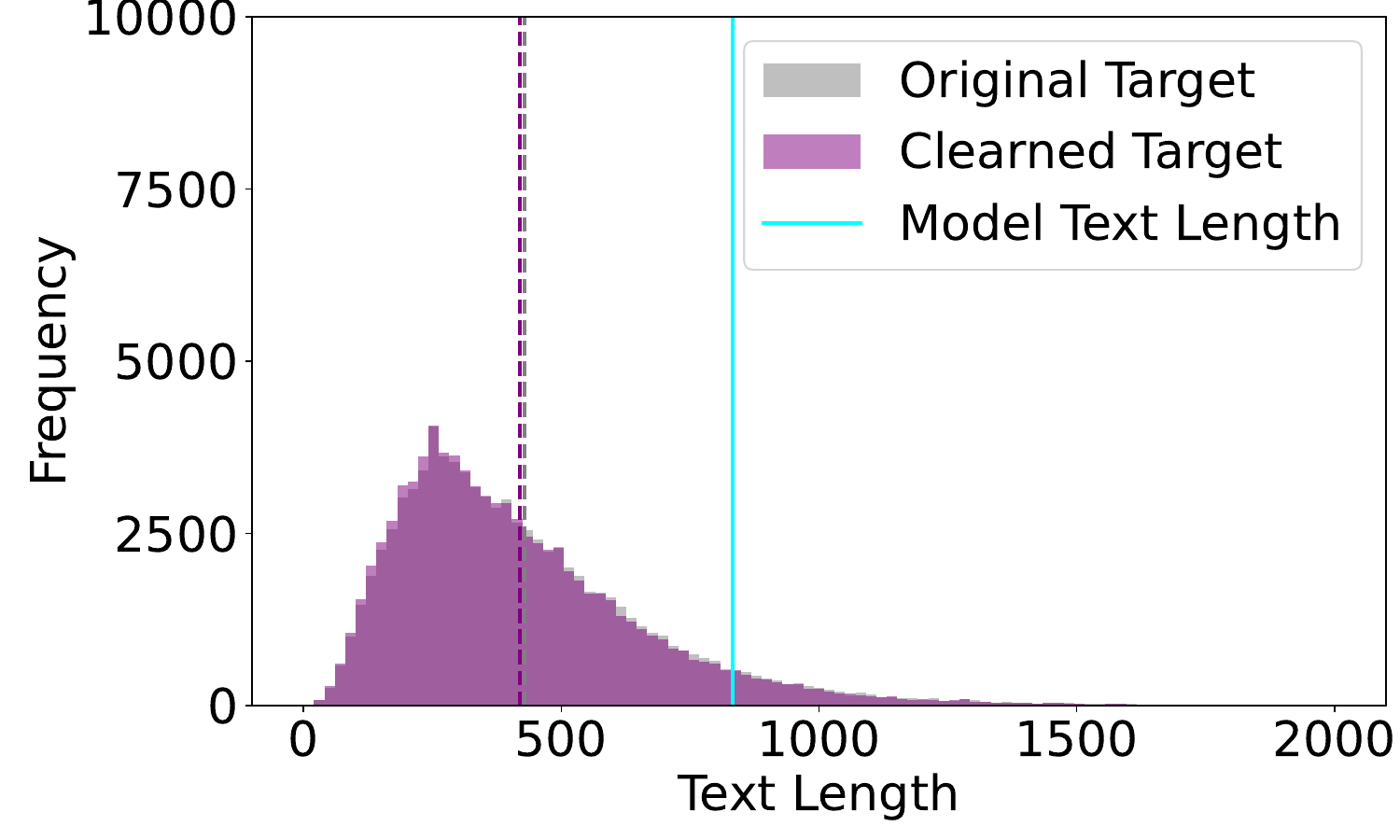}
        \subcaption{{\color{blue}Brief Hospital Course}}
        \label{fig:target-length-brief}
    \end{minipage}\hfill
    \begin{minipage}{0.5\linewidth}
        \vspace*{5mm}
        \centering
        \includegraphics[width=\linewidth]{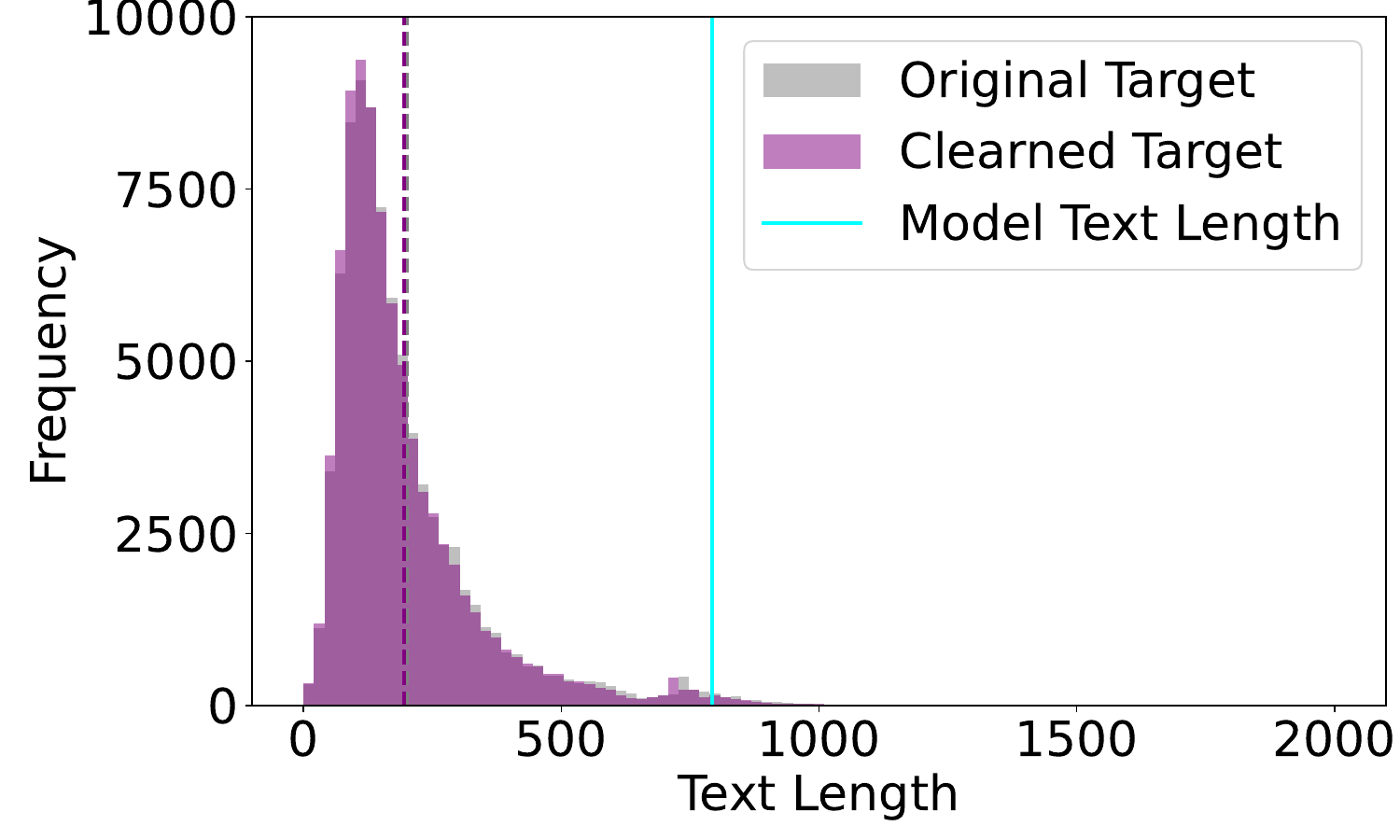}
        \subcaption{{\color{orange}Discharge Instructions}}
        \label{fig:target-lenght-instructs}
    \end{minipage}
    \caption{
Histograms of the text length (in tokens) of the target texts before and after preprocessing for the training and validation sets. The dashed line is the mean. The maximum text length is 832 tokens for ``{\color{blue}Brief Hospital Course}'' and 792 tokens for ``{\color{orange}Discharge Instructions}'', see Table~\ref{tab:model-text-length} in Appendix~\ref{app:fine-tuning} for more details.
}
    \label{fig:stats-target}
\end{figure*}

\begin{table*}[t!]
\centering
\begin{tabular}{lcccc}
\toprule
& \multicolumn{2}{c}{{\color{blue}Brief Hospital Course}} & \multicolumn{2}{c}{{\color{orange}Discharge instructions}}\\
& Original Target & Cleaned Target & Original Target & Cleaned Target \\
\midrule
Min & 2 & 1 & 10 & 10 \\
Max & 4614 & 4452 & 5025 & 4861\\
Mean & 428 & 419 & 201& 195  \\
\bottomrule
\end{tabular}
\caption{Statistical information (in tokens) for histograms in Fig.~\ref{fig:stats-target}.}
\label{tab:stats-target}
\end{table*}

\begin{table*}[t!]
\centering
\begin{tabular}{lcc}
\toprule
& {\color{blue}Brief Hospital Course} & {\color{orange}Discharge instructions}\\
\midrule
Input Text& \multicolumn{2}{c}{1596} \\
\midrule
Generated Text & 832 & 792 \\
Text & 832 & 792 \\
\bottomrule
\end{tabular}
\caption{Maximum text length (tokens).}
\label{tab:model-text-length}
\end{table*}

\begin{table}[t!]
\centering
\begin{tabular}{lc}
\toprule
Batch size & 2\\
Epochs & 4 \\
Learning rate & 1e-4 \\
Precision setting & FP16 \\
Weight decay & 0.01 \\
\bottomrule
\end{tabular}
\caption{Hyperparameters for fine-tuning. }
\label{tab:params}
\end{table}

\begin{table}[t!]
\centering
\begin{tabular}{lc}
\toprule
Dropout probability & 0.05\\
Rank & 4 \\
Target modules & Query \& Value\\
$\alpha$ & 16 \\
\bottomrule
\end{tabular}
\caption{Hyperparameters for LoRA.}
\label{tab:lora-params}
\end{table}

\begin{table}[t!]
\centering
\begin{tabular}{lc}
\toprule
Min length & 10\\
Num beams & 4 \\
Do sample & True \\
Length penalty & 1.1 \\
No repeat $n$-gram size & 4 \\
\bottomrule
\end{tabular}
\caption{Hyperparameters to generate each target discharge summary.}
\label{tab:gen-params}
\end{table}

\subsection{Examples of input text}
Tables~\ref{tab:biref-hospital-course-input} and~\ref{tab:discharge-instructions-input} show examples of input text. These examples illustrate that the ClinicalT5-large model is fine-tuned with different input text for each target discharge summary.

\subsection{Statistical information}
Fig.~\ref{fig:stats-input} shows histograms of the text length (in tokens) of the EHR and the input texts for the training and validation sets. Table~\ref{tab:stats-input} shows the statistical information for these histograms. As shown in Fig.~\ref{fig:stats-input} and Table~\ref{tab:stats-input}, the preprocessing significantly reduces the length of the text.

\section{Details of target text}\label{app:clearned-target}
\subsection{Extraction and concatenation of segments}
In the first process of segment extraction, we divide the text into segments based on blank lines and identify the distinct segments. We then remove spaces and line breaks from each segment and discard empty segments to retain only meaningful segments. Multiple consecutive spaces within each segment are replaced by a single space to improve readability. Finally, we reassemble the cleaned segments with line breaks to make them more suitable for training language models.

\subsection{Examples of preprocessed target text}
Tables~\ref{tab:brie-hospital-course-cleaned-target} and~\ref{tab:discharge-instructions-clearned-target} show examples of the target text before and after preprocessing. These examples illustrate that redundant line breaks are removed after preprocessing.

\subsection{Statistical information}
Fig.~\ref{fig:stats-target} shows histograms of the text length (in tokens) of the target texts before and after preprocessing for the training and validation sets. Table~\ref{tab:stats-target} shows the statistical information for these histograms. As shown in Fig.~\ref{fig:stats-target} and Table~\ref{tab:stats-target}, the preprocessing slightly reduces the length of the text.

\section{Details of fine-tuning}\label{app:fine-tuning}
We used Pytorch~\cite{DBLP:conf/nips/PaszkeGMLBCKLGA19} and huggingface transformers~\cite{DBLP:conf/emnlp/WolfDSCDMCRLFDS20} to implement and fine-tune our models.
We also use peft~\cite{peft} for LoRA.

Table~\ref{tab:model-text-length} shows the text length (in tokens) used by our models.
Table~\ref{tab:params} shows the hyperparameters used for fine tuning.
Table~\ref{tab:lora-params} shows the hyperparameters used for LoRA.
Table~\ref{tab:gen-params} shows the hyperparameters to generate each target discharge summary.

\end{document}